\documentclass[letterpaper]{article} 
\usepackage{aaai2026}  
\usepackage{times}  
\usepackage{helvet}  
\usepackage{courier}  
\usepackage[hyphens]{url}  
\usepackage{graphicx} 
\usepackage{natbib}  
\usepackage{caption} 
\usepackage{algorithm}
\usepackage{algorithmic}

\usepackage{newfloat}
\usepackage{listings}
\DeclareCaptionStyle{ruled}{labelfont=normalfont,labelsep=colon,strut=off} 
\floatstyle{ruled}
\newfloat{listing}{tb}{lst}{}
\floatname{listing}{Listing}
\usepackage{subcaption}

\title{The Illusion of Procedural Reasoning: \\Measuring Long-Horizon FSM Execution in LLMs}
\author {
    Mahdi Samiei\textsuperscript{\rm 1},
    Mahdi Mansouri\textsuperscript{\rm 1},
    Mahdieh Soleymani Baghshah\textsuperscript{\rm 1}
}
\affiliations {
    \textsuperscript{\rm 1}Department of Computer Engineering, Sharif university of Technology\\
    mm.samiei@sharif.edu, uni.mahdi.mansouri@gmail.com, soleymani@sharif.edu
}

\usepackage{bibentry}

\begin{document}

\maketitle

\begin{abstract}
Large language models (LLMs) have achieved remarkable results on tasks framed as reasoning problems, yet their true ability to perform procedural reasoning, executing multi-step, rule-based computations remains unclear.
Unlike algorithmic systems, which can deterministically execute long-horizon symbolic procedures, LLMs often degrade under extended reasoning chains, but there is no controlled, interpretable benchmark to isolate and measure this collapse.
We introduce Finite-State Machine (FSM) Execution as a minimal, fully interpretable framework for evaluating the procedural reasoning capacity of LLMs.
In our setup, the model is given an explicit FSM definition and must execute it step-by-step given input actions, maintaining state consistency over multiple turns. This task requires no world knowledge, only faithful application of deterministic transition rules, making it a direct probe of the model’s internal procedural fidelity. We measure both Turn Accuracy (local correctness) and Task Accuracy (global long-horizon correctness) to disentangle immediate computation from cumulative state maintenance.
Empirical results reveal systematic degradation as task horizon or branching complexity increases. Models perform significantly worse when rule retrieval involves high branching factors (many actions per state) than when memory span is long (many states, few actions). Larger models show improved local accuracy but remain brittle under multi-step reasoning unless explicitly prompted to externalize intermediate steps.
These findings expose a consistent illusion of procedural reasoning: LLMs can mimic algorithmic behavior for short traces but fail to sustain coherent execution as procedural depth grows. FSM-based evaluation offers a transparent, complexity-controlled probe for diagnosing this failure mode and guiding the design of inductive biases, memory mechanisms, and reasoning scaffolds that enable genuine long-horizon procedural competence.
By grounding “reasoning” in measurable execution fidelity rather than surface correctness, this work helps establish a rigorous experimental foundation for understanding and improving the algorithmic reliability of LLMs.
\end{abstract}

\begin{figure*}[t]
\centering
\begin{subfigure}[t]{0.48\textwidth}
\centering
\includegraphics[width=\linewidth]{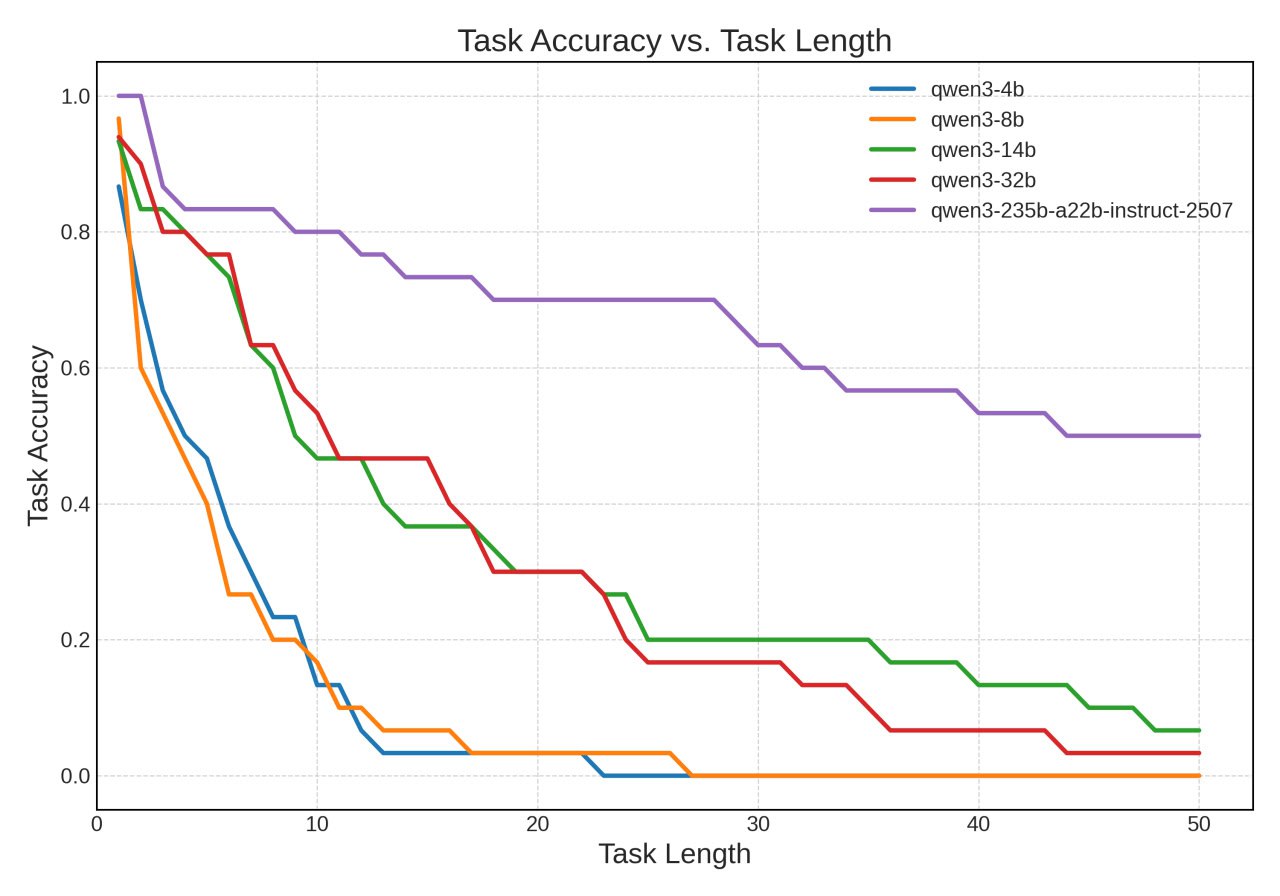}
\caption{\textbf{Task Accuracy vs. Model Scale.} While scaling improves cumulative execution fidelity, even the largest model (Qwen3-235B) reaches only about 50\% overall task accuracy, indicating persistent long-horizon degradation.}
\label{fig1}
\end{subfigure}\hfill
\begin{subfigure}[t]{0.48\textwidth}
\centering
\includegraphics[width=\linewidth]{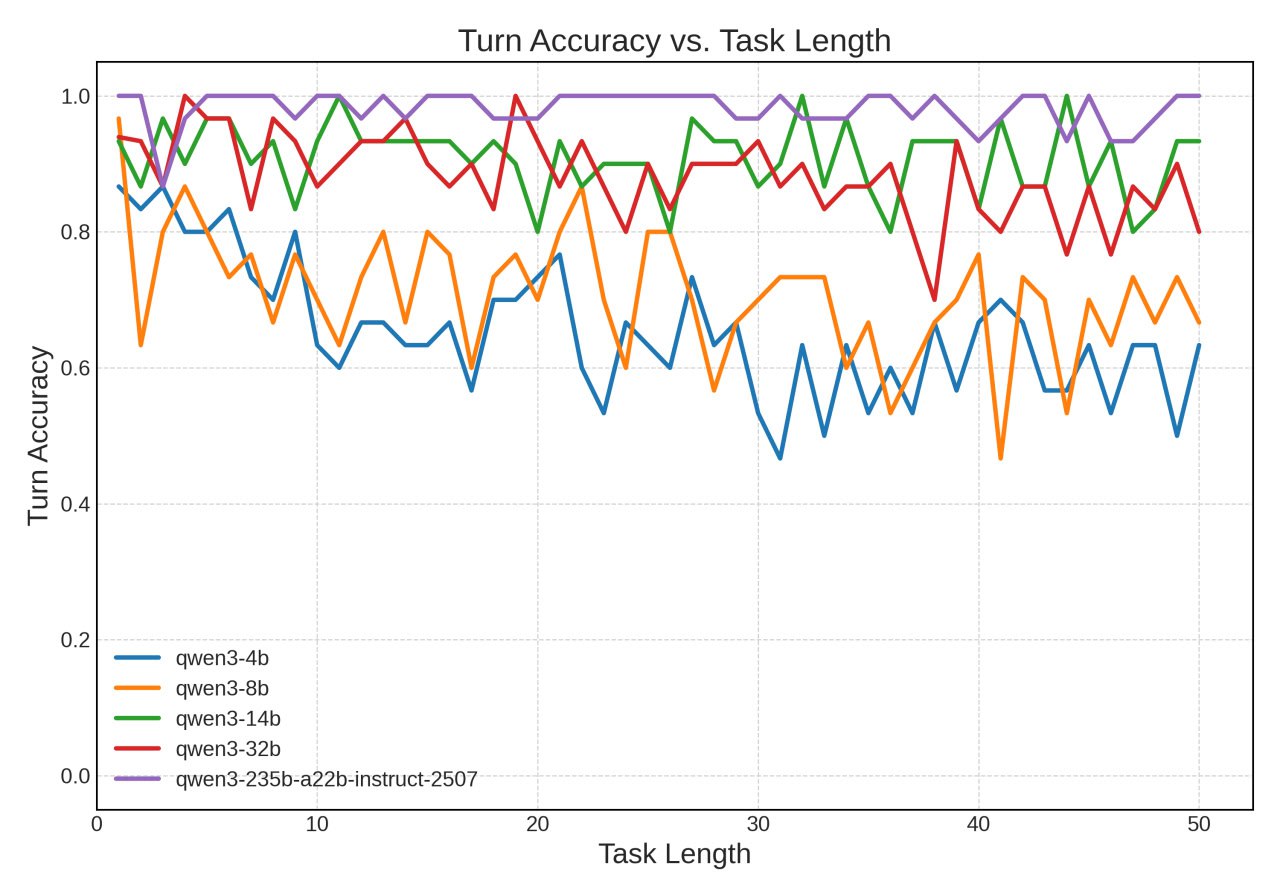}
\caption{\textbf{Turn Accuracy vs. Model Scale.} Larger Qwen models achieve higher per-turn correctness when executing FSM transitions. Accuracy increases steadily with parameter count, suggesting stronger local rule adherence in larger models.}
\label{fig2}
\end{subfigure}
\caption{Task accuracy and Turn accuracy comparison for different models}
\label{fig:combined}
\end{figure*}

\section{Introduction}
LLMs have demonstrated striking performance on a wide range of benchmarks that are often described as requiring reasoning. Yet a central question remains unsettled: do LLMs possess genuine reasoning competence, or are they primarily leveraging statistical pattern matching over surface forms and short-horizon reasoning problems? Recent studies have cast doubt on the former view by showing that LLM performance frequently collapses as samples complexity increases, measured by longer compositions, deeper dependency chains, or more distractors and that apparent successes are often concentrated on lower complexity instances that may be susceptible to template memorization or training data contamination
\citep{shojaee2025illusionthinkingunderstandingstrengths, mirzadeh2025gsmsymbolic, sun2025omegallmsreasonoutside, sinha2025illusiondiminishingreturnsmeasuring, paqaleh2025bridgingreasoninglearningunmasking, zhou2025gsminfinitellmsbehaveinfinitely}
. These observations have sharpened the debate over whether current LLMs truly reason or merely extrapolate familiar patterns.

A growing line of work proposes decomposing reasoning into two phases: planning and execution
\cite{shojaee2025illusionthinkingunderstandingstrengths, sinha2025illusiondiminishingreturnsmeasuring}
. Planning concerns deriving a solution strategy or algorithm for a problem; execution concerns faithfully carrying out that strategy step by step. Empirically, even when the algorithmic plan is supplied to the model, performance can still degrade precipitously as instance complexity rises, implicating failures in the execution phase rather than in plan discovery per sample.

\citet{sinha2025illusiondiminishingreturnsmeasuring} provide complementary evidence from purely executional tasks in which the "plan" reduces to simple retrieval and aggregation—for example, retrieving numeric values by given word keys and summing them, yet models struggle to generalize beyond short-horizon templates. This work also highlight a self-conditioning effect whereby incorrect intermediate outputs feed back into the context, increasing the likelihood of subsequent errors and inducing error cascades over multi-step interactions.

Despite substantial progress, existing benchmarks rarely disentangle planning from execution under precisely controlled complexity, nor do they provide a minimal environment in which the correctness of each intermediate step is unambiguous. Building on these perspectives, we propose a complementary line of inquiry: FSMs as a controllable probe of reasoning and execution complexity in LLMs. FSMs represent explicit, interpretable procedural structures with precisely quantifiable complexity—defined by the number of states, transitions, and branching dependencies. When an LLM is tasked with executing an FSM—simulating the state updates for a given input sequence from an initial state until a goal condition is met—it must perform a sequence of discrete, deterministic reasoning steps. Unlike symbolic puzzles or text-based math problems, FSMs eliminate ambiguity in what constitutes a correct step, thereby enabling precise measurement of systematic reasoning failures as structural complexity and horizon length increase.

Our experiments show that even bare-bones FSM execution is challenging for current LLMs. Accuracy declines predictably with the number of actions and input length, and we observe clear self-conditioning dynamics: an early state-tracking mistake propagates through subsequent steps and amplifies downstream error rates. Moreover, enabling "thinking" prompting does not resolve the collapse at higher complexities, suggesting that the bottleneck lies in reliable stepwise execution rather than plan articulation. FSM-based evaluation thus offers a minimal, interpretable environment to quantify this failure mode and to ground future research on inductive biases, modular controllers, training curricula, and external memory mechanisms that may enable true generalization to increasing procedural complexity.

\begin{figure*}[t]
\centering
\begin{subfigure}[t]{0.48\textwidth}
\centering
\includegraphics[width=\linewidth]{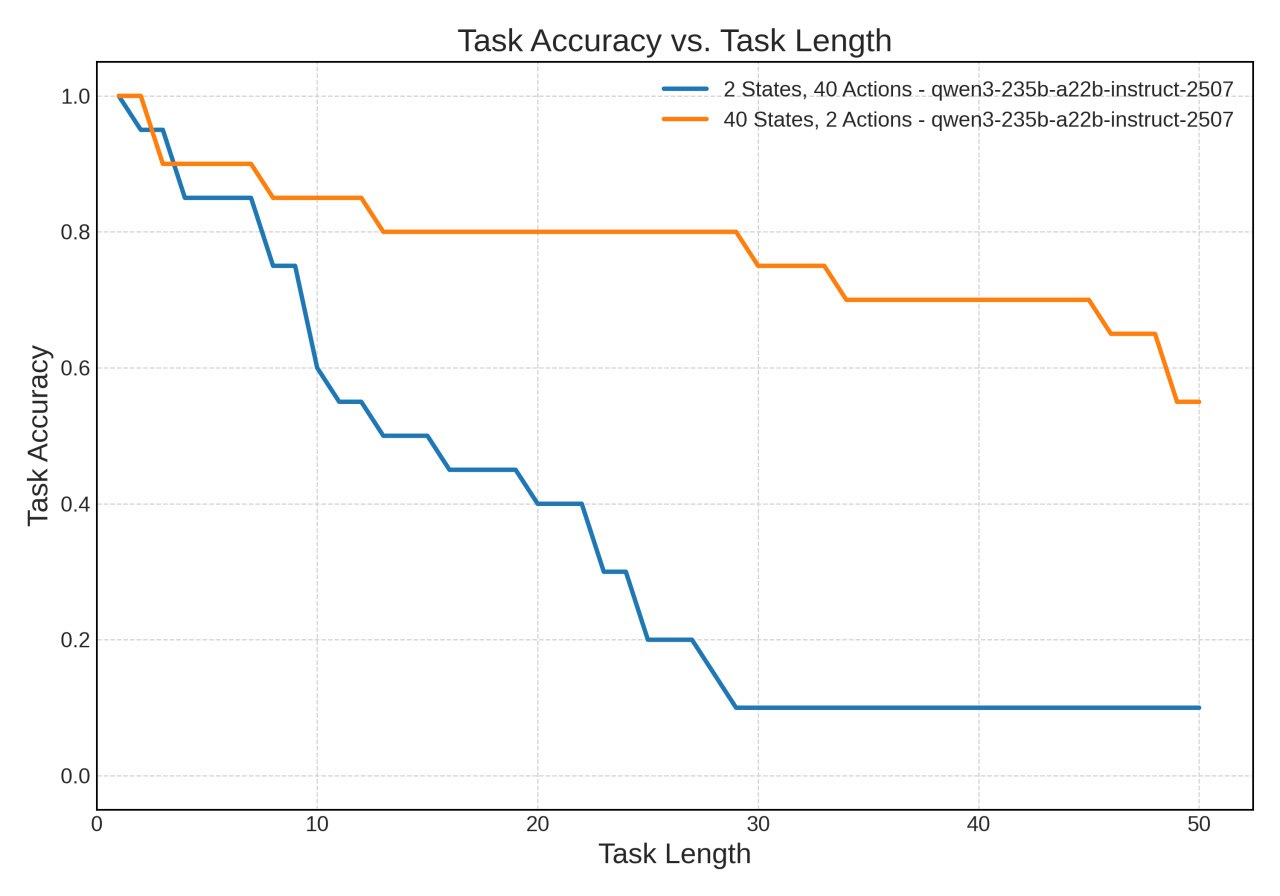}
\caption{Task accuracy comparison for a 2-state/40-action vs 40-state/2-action setup. The setup with 2 states and 40 actions are much harder to solve for an LLM while using 40 states and 2 actions will lead to much better accuracy for long-horizon tasks.}
\label{fig1}
\end{subfigure}\hfill
\begin{subfigure}[t]{0.48\textwidth}
\centering
\includegraphics[width=\linewidth]{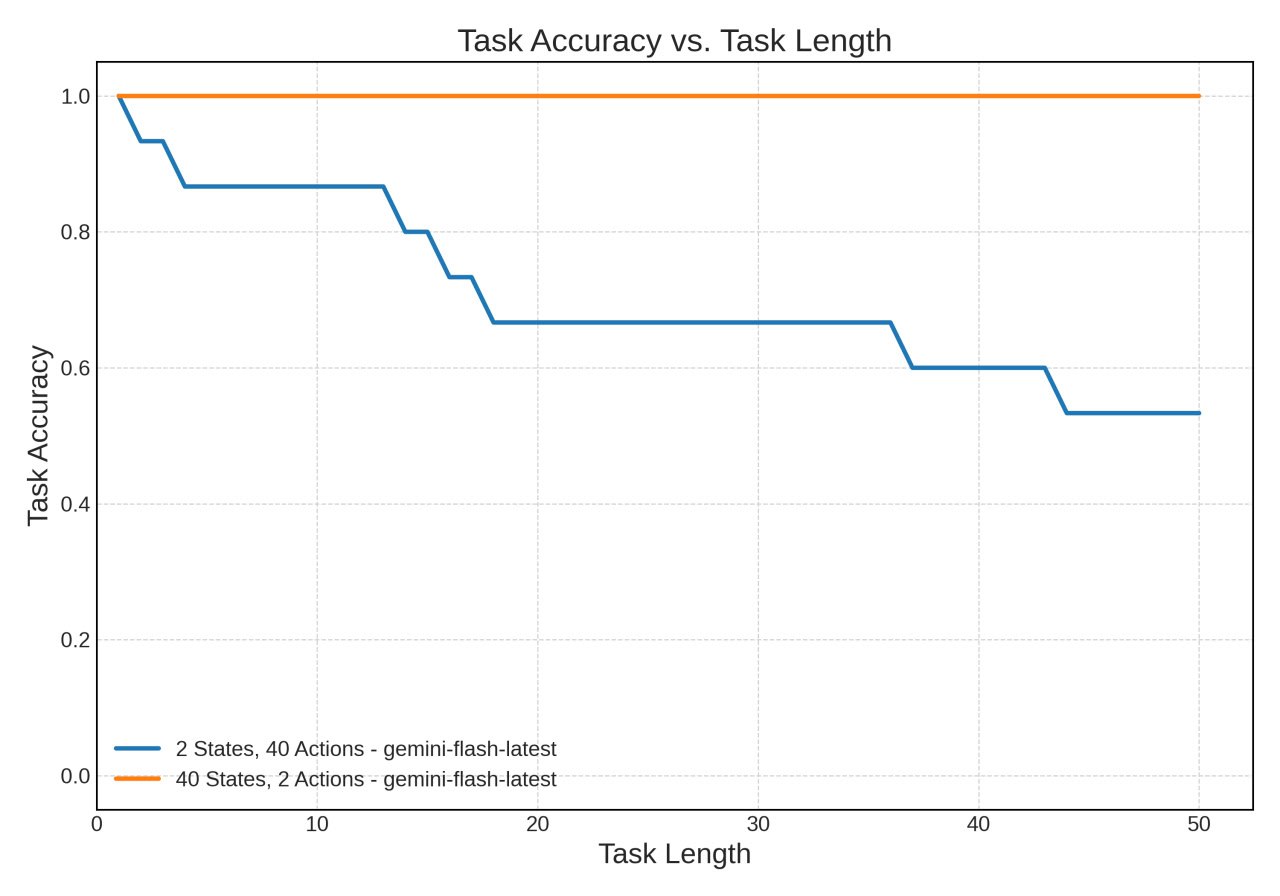}
\caption{Task accuracy comparison for a 2-state/40-action vs 40-state/2-action setup. Gemini-2.5-flash can reach only about 50\% accuracy at the end of a task with 50 steps with only 2 states and 40 actions meanwhile this model can reach 100\% accuracy for a 40-state/2-action setup.}
\label{fig2}
\end{subfigure}
\caption{Task accuracy comparison for a Wide \& Shallow setup vs Deep \& Narrow setup.}
\label{fig:combined}
\end{figure*}

\section{Formulation}




To completely evaluate an LLM's ability to maintain long-term, procedural state, we employ the FSM as a ground-truth framework. An FSM is a computational model that is always in one of a finite set of states. Its 'next' state is determined only by its 'current' state and a given 'action,' according to a predefined set of transition rules. This makes it an ideal tool for this experiment: it is computationally simple for a classical machine but challenging for an LLM, as it requires perfect adherence to arbitrary rules and flawless state memory over long interactions, rather than relying on semantic or world knowledge.

Formally, we define our FSM as a 4-tuple $M = (Q, \Sigma, \delta, q_0)$, where:
\begin{itemize}
    \item $Q$ is a finite set of states.
    \item $\Sigma$ is a finite set of actions.
    \item $\delta: Q \times \Sigma \rightarrow Q$ is the transition function that maps a (state, action) pair to a new state.
    \item $q_0 \in Q$ is the designated initial state.
\end{itemize}

To ensure the task tests rule-following rather than semantic inference, we populate the states ($Q$) and actions ($\Sigma$) from disjoint sets of simple, single-token English words (e.g., states as nouns like 'cat', 'desk'; actions as adjectives like 'red', 'fast'). This prevents the model from "guessing" transitions based on word association.

The complete FSM definition—including all states, actions, the initial state $q_0$, and a full list of transitions (e.g., \texttt{From cat, on action red, go to desk.})—is provided to the LLM in a system prompt. This prompt sets the LLM's role as an FSM executor and strictly defines the output format as \texttt{<state>FinalState</state>}. At each conversational turn, the model receives a sequence of actions and outputs the resulting state, which becomes the starting state for the next turn.

To evaluate performance, we define two metrics:
\begin{enumerate}
    \item \textbf{Turn Accuracy:} This metric evaluates the model's computational correctness on a turn-by-turn basis. It checks if the model's reported state at turn $t$ is the correct result after applying the action sequence given in the prompt from turn $t-1$. Turn Accuracy answers the question: "Did the model perform this single calculation correctly, started from the previous state even if it was wrong?"

    \item \textbf{Task Accuracy:} This is a much stricter metric measuring long-term fidelity. It checks if the model's reported state at turn $t$ is the true state, assuming a perfect path from the initial state $q_0$. Task accuracy is 1 only if every preceding turn (including the current one) was also correct. Task Accuracy answers the question: "Has the model remained on the correct path since the very beginning?"
\end{enumerate}

The distinction is critical: Turn accuracy measures the model's immediate processing ability, while task accuracy measures its long-term state-holding capacity. A model can have a high turn accuracy (it calculates correctly from its own, flawed state) but a low task accuracy (it was knocked off the correct path many turns ago and never recovered).

\section{Experiments}

We applied our FSM formulation to evaluate several LLMs. Our experiments were designed to investigate three primary questions: (1) the effect of model scale on state-tracking fidelity, (2) the impact of FSM structure (state-space vs. action-space complexity), and (3) the model's ability to handle multi-step sequential instructions within a single turn.

\subsection*{Scaling Effects on Rule Adherence}

We first investigated the effect of model scale on a baseline FSM task (4 states, 5 actions). As shown in Figure \ref{fig1}, we tested various Qwen models of different parameter sizes. The results indicate a strong positive correlation between model scale and both \textbf{Turn Accuracy} and \textbf{Task Accuracy}. This suggests that larger models possess superior capabilities for strict rule-adherence and long-term state maintenance, with Qwen3-235b substantially outperforming all other models in this test. Even though this outperforming model could get about 50\% accuracy in the tasks.

A key finding, visible in Figure \ref{fig1}, is the effect of "negative self-conditioning," particularly in smaller models. These models exhibit a clear decaying pattern in Turn Accuracy over time.

This is a significant observation. While Turn Accuracy is measured relative to the model's own previous state—and should, in theory, remain independent of past task failures—the data suggests it is not. This pattern, which echoes the findings of Sinha et al. (2025), implies that as a model makes mistakes, its ability to perform the next computational step correctly (even from its own flawed state) is also degraded.

Therefore, for these models, Turn Accuracy is not truly independent of the conversational history. It appears that as the context window becomes "polluted" with the model's own errors, its immediate processing ability is compromised, creating a cycle of compounding failure.

\begin{figure*}[t]
\centering
\includegraphics[width=0.5\textwidth]{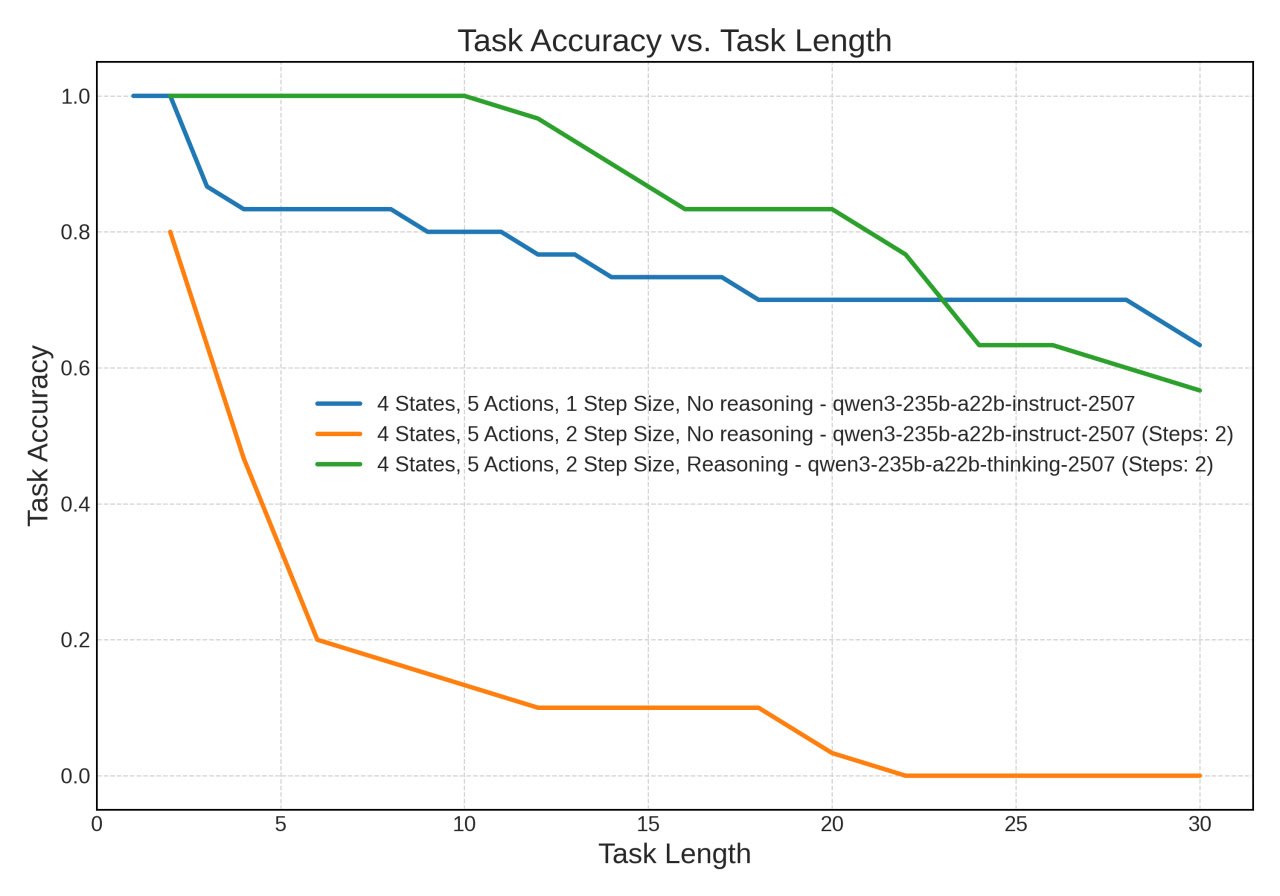} 
\caption{Increasing step size to 2 will result a huge performance degradation on a 4-state/5-action setup. It indicated that steps should also be atomized to reach high task accuracy. Using reasoning in this setup will lead to much higher performance in the cost of reasoning tokens generated by the model.}
\label{fig2}
\end{figure*}

\subsection*{State-Space vs. Action-Space: A Counter-Intuitive Finding}

A surprising and central finding of our work is illustrated in Figure 2. We tested two high-performing models (Qwen3-235b and Gemini-2.5-flash) on two FSM configurations with an identical number of total transitions (80):
\begin{itemize}
    \item \textbf{Config 1 (Wide \& Shallow):} 2 states, 40 actions.
    \item \textbf{Config 2 (Deep \& Narrow):} 40 states, 2 actions.
\end{itemize}

Intuition might suggest that Config 1 (fewer states) would be easier, as even a random guess of the final state would have a 50\% probability of being correct.

Our findings demonstrate the opposite: both models performed significantly better on the FSM with 40 states and 2 actions.

We hypothesize this is not a failure of \textit{state memory} but a failure of \textit{rule retrieval}. In the 2-state/40-action FSM, each state has a very high \textit{branching factor}. When processing a turn, the model must locate the single correct transition rule from a list of 40 very similar rules (e.g., \texttt{From state\_A, on action\_1...}, \texttt{From state\_A, on action\_2...}). This "needle in a haystack" problem appears to confuse the model's attention mechanism, leading to frequent errors. Conversely, the 40-state/2-action FSM has a low branching factor (2 rules per state), making rule retrieval trivial, even if the total \textit{state space} is large.

This leads to a practical design principle: when building LLM-based systems (like multi-agent frameworks or Chain-of-Thought prompts), it is preferable to design workflows with many simple decision points (many states, few actions) rather than few complex decision points (few states, many actions).

\subsection*{Instruction Complexity and Mitigation via Reasoning}

Finally, we investigated the impact of \textit{instruction complexity} per turn, as shown in Figure 3. In all previous experiments, we used a step size of one (one action per turn). In this experiment, we also tested a step size of two (two sequential actions in a single prompt).

As the plot illustrates, performance degraded completely when models were asked to process two sequential actions in one turn. This highlights a critical limitation: LLMs struggle to perform multiple, sequential computational steps "in-head" within a single generation. It is far more effective to break the process into distinct user prompts.

Crucially, we found this degradation can be significantly mitigated by enabling model reasoning. When the model was prompted to "think" through the two steps, it effectively used its context window as a scratchpad to record the \textit{intermediate state} after the first action, before proceeding to the second. This externalization of the intermediate step allowed it to compute the final state correctly. The results were much better, though this came at the cost of additional reasoning tokens.

\section{Conclusion}
This work introduced FSMs as a minimal yet powerful framework for probing the procedural reasoning capabilities of large language models. By decoupling symbolic execution from semantic interpretation, FSM-based evaluation allows precise measurement of a model’s ability to maintain and manipulate internal state over long horizons. Across a range of models and configurations, we observed a consistent pattern: LLMs can follow short procedural traces accurately but fail predictably as the complexity or temporal depth of the task increases. The degradation follows a self-conditioning dynamic, where early local errors propagate through subsequent steps, leading to global collapse in long-term state fidelity.

Our experiments reveal several key insights. First, model scale improves short-horizon execution but does not fundamentally eliminate long-horizon instability, indicating that scaling alone may not yield genuine procedural competence. Second, the sharp performance asymmetry between wide and deep FSMs highlights that failures often stem from rule retrieval under high branching factors rather than from limited memory over large state spaces. Finally, the catastrophic drop in accuracy for multi-step (two-action) instructions demonstrates the fragility of internal, unexternalized reasoning, an effect that can be partially mitigated when the model is encouraged to externalize intermediate steps through explicit reasoning or scratchpad generation.

Together, these findings suggest that current LLMs exhibit an illusion of procedural reasoning: they can imitate the appearance of algorithmic control in short contexts but lack stable internal mechanisms for multi-step execution. FSM-based evaluation offers a simple, interpretable testbed for isolating this failure mode and guiding the development of architectures and prompting strategies that incorporate compositional inductive bias, explicit memory, or modular execution scaffolds. Future research should explore integrating symbolic controllers, structured reasoning traces, or recurrent supervision to endow models with the ability to sustain coherent, rule-based computation across extended horizons, an essential step toward genuine systematic reasoning.

\bibliography{aaai2026}

\end{document}